# Extremal GloVe : Theoretically Accurate Distributed Word Embedding by Tail Inference

Hao, Wang*

Ratidar.com

Haow85@live.com

Beijing, China

Distributed word embeddings such as Word2Vec and GloVe have been widely adopted in industrial context settings. Major technical applications of GloVe include recommender systems and natural language processing. The fundamental theory behind GloVe relies on the selection of a weighting function in the weighted least squres formulation that computes the powered ratio of word occurrence count and the maximum word count in the corpus. However, the initial formulation of GloVe is not theoretically sound in two aspects, namely the selection of the weighting function and its power exponent is ad-hoc. In this paper, we utilize the theory of extreme value analysis and propose a theoretically accurate version of GloVe. By reformulating the weighted least squares loss function as the expected loss function and accurately choosing the power exponent, we create a theoretically accurate version of GloVe. We demonstrate the competitiveness of our algorithm and show that the initial formulation of GloVe with the suggested optimal parameter can be viewed as a special case of our paradigm.

**Additional Keywords and Phrases:** GloVe, distributed word embedding, tail inference, extreme value analysis

## 1 Introduction

Vector representation of human words is a major breakthrough in the history of natural language processing. The first invention in the field is Word2Vec [1], within a couple of years, there emerges voluminous research on the topic, ranging from distributed word representation to distributed document representation. The significance of the invention can never be over-emphasized because the dense representation of human words is highly useful in many application scenarios such as recommender system and chatbots.

The distributed word embedding algorithm GloVe [2] was invented in the year of 2014 and has had a comparable reputation as Word2Vec. It is among the must-read materials of many higher education institutes. The GloVe model has 2 parameter selection steps, namely the weighting function of the weighted least squares formulation and the power exponent of the weighting function. Both parameters were chosen heuristically in the initial formulation of the GloVe model, which leaves space for improvement in the theoretical domain.

Statisticians have an enduring interest in linguistics. Zipf Law was invented decades ago to capture the power law distribution of the word occurrence in the document corpus. The initial formulation of GloVe uses Zipf distribution to model the weighting function of the weighted least squares formulation. However, this is ad-hoc choice and not theoretically justified. The selection of the power exponent of the initial formulation of GloVe is also ad-hoc. A commonly suggested choice is 0.75 when the dimension is 100, which lacks theoretical support.

---

\* Place the footnote text for the author (if applicable) here.

In this paper, we reformulate the loss function of GloVe as the expected squared error. We model the weighting function as the joint distribution of 2 different words, which could be simplified as the product of 2 Zipf distribution if word occurences are i.i.d. distributed. As for the power exponent of the Zipf distribution, we take advantage of extreme value analysis theory to estimate its value. We illustrate the theoretical soundness of our approach in the following sections and prove in the experimental section that our method yields competitive performance.

## 2 related work

Distributed word embeddings ([1][2]) is a hot research topic since the invention of Word2Vec ([1]). There are many research papers on how to improve the word embedding ([1][2]), sentence embedding ([3][4]) and paragraph embedding ([5]). As we all know, distributed word embedding is widely applicable in web mining tasks, the theoretical accuracy and technical soundness behind the models is a critical issue in the industry.

The problem formulation of GloVe is a weighted least squares problem that uses Zipf distribution to model the weighting function. Tail inference techniques such as Pickands Estimator ([6]), Hill Estimator ([7]), and QQ Estimator ([8]), among dozens of similar inventions are very well suited for the task of parameter estimation for the Zipf distribution. In this paper, we reformulate the weighted least squares loss function as an expected loss error function and use tail inference for parameter estimation.

In the year of 2018, two new techniques are invented, namely elmo [9] and BERT [10]. Both technologies are important in the field of distributed word embeddings, and in the larger picture, natural language processing. However, unlike GloVe, both technologies are deep learning models that requires more powerful computing facilities and training time. Therefore, early distributed word embedding models such as word2vec and GloVe are still applicable in real world commercial environments.

## 3 Tail Inference

The initial formulation of GloVe loss function is as follows:

$$J = \sum_{i,j=1}^{V} f\left(X_{ij}\right) \left(w_i^T \bar{w}_j + b_i + \bar{b}_j - \log X_{ij}\right)^2$$

The author of GloVe models the weighting funciton f with Zipf distribution with an officially suggested power exponent of 0.75 on both the initial paper and the Stanford NLP's GitHub page. The choice of weighting function and parameter is ad-hoc. Although the authors of GloVe resorted to the theory of kernel smoothing and stated that the choice of the weighting function only needs to satisfy certain conditions, but they did not specify what the optimal or legitimate choice is for the formulation. Please notice although the formulation bears resemblance to the kernel smoothing formulation, there are strikingly huge differences. For example, there is no kernel bandwidth defined in the formulation, and the official specification of the weighting function does not really make it the right kernel smoothing formulation. Or, to be more precise, the relation of distances between variables are defined in a highly complex manner than they appear in the weighting function if the formulation is to be interpreted in a rigor kernel smoothing context.

In this paper, we reformulate the weighted least squares loss function as the expected loss function, where the weighting function f is considered as the joint distribution function of the occurrence of word i and word j, namely :

$$J = \sum_{i,j=1}^{n} E\left[\left(w_i^T w_j + b_i + b_j - \log X_{i,j}\right)^2\right]$$

To compute the expected loss error, we need to specify the probability distribution of $X_i$ and $X_j$. We compute the expected loss error using the following formula :

$$J = \sum_{i,j=1}^{n} E\left[\left(w_i^T w_j + b_i + b_j - \log X_{i,j}\right)^2 | X_i, X_j\right] \times f\left(X_i, X_j\right)$$

, where E is the conditional expected loss function and f is the joint probability distribution of word pairs.

We assume the distribution of the word occurences are independently identically distributed, namely, we can decompose the joint probability distribution as the product of 2 zipf distributions as specified by the initial formulation of the GloVe model (the initial model contains only 1 zipf distribution) :

$$f(x) = \begin{cases} (x/x_{max})^\alpha & \text{if } x < x_{max} \\ 1 & \text{otherwise} \end{cases}$$

, our loss function now becomes :

$$J = \sum_{i,j=1}^{n} \left(w_i^T w_j + b_i + b_j - \log X_{i,j}\right)^2 \times \left(\frac{X_i}{\max(X_i)}\right)^\alpha \times \left(\frac{X_j}{\max(X_j)}\right)^\alpha$$

Please notice the resemblance of the new formulation with the initial definition of Glove. It only requires a minimal amount of modification to the Stanford NLP code to make the new formulation work. The reason why we choose the product of Zipf functions as the joint probability function is it is well known that the probability distribution of single words in language corpus follows Zipf law.

The remaining question is how to select the power exponent of the Zipf distribution, namely $\alpha$. In this paper, we propose to use tail inference to approximate the power exponent of the Zipf distribution. Tail inference is a technique of the extreme value analysis. It uses order statistics to approximate the power exponent of Zipf distribution. Using tail inference to estimate the power exponent of Zipf distribution is a standard practice in financial sector. We omit the discussion and proofs behind the tail index inference theory in our paper and leave the readers to extreme value analysis textbooks for reference and study.

There are many tail index estimators for approximation of the power exponent in theory and we choose the following estimators to test our algorithm :

1. Pickands Estimator ([6]):

$$p = \frac{1}{\ln 2} \ln\left(\frac{X_{M:n} - X_{2M:n}}{X_{2M:n} - X_{4M:n}}\right)$$

2. Hill Estimator ([7]):

$$p = \frac{1}{k} \sum_{i=1}^{k} \ln X_{i:n} - \ln X_{k+1:n}$$

3. Adapted Hill Estimator ([7]) :

$$p = \frac{1}{k} \sum_{i=1}^{k} \ln(UH_i) - \ln(UH_{k+1})$$

where :

$$UH_i = X_{(i+1):n}\left(\frac{1}{i}\sum_{j=1}^{i} \ln X_{j:n} - \ln X_{(i+1):n}\right)$$

4. Moment Estimator ([11]):

$$p = M_1 + 1 - \frac{1}{2}\left(1 - \frac{M_1^2}{M_2}\right)^{-1}$$

where :

$$M_j = \frac{1}{k} \sum_{i=1}^{k} \left(\ln X_{i:n} - \ln X_{(k+1):n}\right)^j$$

5. QQ Estimator ([8]):

$$p = \frac{\sum_{i=1}^{k} \ln \frac{i}{k+1} \left\{ \sum_{j=1}^{k} \ln X_{j:n} - k \ln X_{i:n} \right\}}{k \sum_{i=1}^{k} \left( \ln \frac{i}{k+1} \right)^2 - \left( \sum_{i=1}^{k} \ln \frac{i}{k+1} \right)^2}$$

6. Peng's Estimator ([12]):

$$p = \frac{M_2}{2M_1} + 1 - \frac{1}{2} \left( 1 - \frac{M_1^2}{M_2} \right)^{-1}$$

where :

$$M_i = \frac{1}{k} \sum_{j=1}^{k} \left( \ln X_{j:n} - \ln X_{(k+1):n} \right)^i$$

In the formulas of these estimators, $X_{i:n}$ means the i-th largest order statistic of the word occurrences of the corpus. We compute the ranks of words in the training corpus and plug them into the formulas. Then we use these estimators to approximate the power exponent of the probability distribution function.

Our theory may seem wrong at the first sight, because we are using the power exponent of ranks to substitute the power exponent of frequencies in the initial formulation. But our theory is actually solid, because the expected loss error uses joint distribution of words as the probability function, which is actually the product of the frequencies of 2 words (similar to the initial formulation). As specified in Zipf's law, the frequency is the powered inverse of the ranks. So the power exponent of ranks in place of the initial power exponent is correct by simple calculation and mathematical thinking.

By reformulating the expected loss function and accurate computation of the Zipf distribution parameter, we provide a much more accurate and solid foundation for the theory of GloVe model. In the next section, we show by experiments that our approach produces comparably favorable results compared with the initial formulation of the GloVe model.

Please notice the initial formulation of GloVe is actually a special case of our Extremal Glove model. If we set the power exponent of the Zipf distribution to be 0.75/2, we obtain the classic GloVe formulation with the suggested optimal power exponent.

## 4 Experiment

We modify the source code of the GloVe model in the Stanford NLP package and test our new model using the test data coming with the official GloVe GitHub repository. We illustrate the experimental results as follows (vector dimension = 50, maximum number of training iteration = 15):

|   | GloVe | Pickands | Hill | Adaptive Hill | Moment | QQ | Peng |
|---|---|---|---|---|---|---|---|
| Semantic Accuracy | 28.01% | 5.58% | 21.48% | 12.46% | 11.35% | 20.06% | 4.58% |
| Syntactic Accuracy | 20.24% | 3.42% | 11.26% | 5.61% | 5.95% | 20.27% | 2.82% |
| Total Accuracy | 23.47% | 4.32% | 15.51% | 8.46% | 8.20% | 20.18% | 3.56% |

The size of the vocabulary file size is 71290. The parameter setting for the vanilla GloVe model is chosen as shown on the official paper and Github source code page, namely the weighting function is Zipf distribution with the cut-off threshold setting to 100 and the power exponent being 0.75. From the experiments, we know that QQ estimator yields the best performance among the tail inference estimators when the parameter size is set to be 1/5 of the vocabulary size. The result is a bit poorer than the officially suggested ad-hoc heuristic in Semantic Accuracy, but better in the Syntactic Accuracy, and the theory behind our Extremal GloVe model is much more accurate in theoretical domain. Therefore we suggest to use QQ Estimator in Extremal GloVe model. In addition, the initial formulation of GloVe with power exponent 0.75 is just a special case of our new formulation.

# 5 conclusion

In this paper, we propose a new version of GloVe model called Extremal GloVe. We suggest reformulating the initial weighted least squares loss function as the expected loss function so tail inference could be applied to estimate the parameters of the weighting distribution. In this way, the whole theory behind GloVe is more accurate and the performance is comparable with the officially suggested optimal parameter selection. We prove that the initial formulation of the GloVe model with the suggested optimal choice of parameter is just a special case of our proposed paradigm.

In future work, we would like to explore further in this direction, so we could find the optimal weighting function and parameters that are not only theoretically accurate but also improves the accuracy of the model. We would also like to explore other formulation of the GloVe model so we could achieve better performance with accurate theoretical support.